\documentclass[aoas,MSNbibl,nameyear,dvips]{arximspdf}
\usepackage{dcolumn}
\usepackage{multirow}
\usepackage{stfloats}
\usepackage{graphicx}

\doi{10.1214/12-AOAS625} 
\volume{7}
\issue{2}
\pubyear{2013}
\firstpage{714}
\lastpage{738}

\makeatletter

\newcolumntype{d}[1]{D{.}{.}{#1}}

\fnbelowfloat

\newcommand{\EV}{\mathbb E}

\makeatother

\begin{document}
\begin{frontmatter}

\title{Object-oriented Bayesian networks for a decision support system for
antitrust enforcement\thanksref{T1}}
\runtitle{Decision support system}

\thankstext{T1}{Supported by a PRIN07 Italian Research Grant.
Sections \protect\ref{secCP}, \protect\ref{secPD} and \protect\ref{secRepeatedPD}
are due to C. Vergari, the remaining sections are due to J. Mortera and
P. Vicard.}

\begin{aug}
\author[A]{\fnms{Julia} \snm{Mortera}\corref{}\ead[label=e1]{mortera@uniroma3.it}},
\author[A]{\fnms{Paola} \snm{Vicard}\ead[label=e2]{vicard@uniroma3.it}}
\and
\author[B]{\fnms{Cecilia} \snm{Vergari}\ead[label=e3]{cecilia.vergari@unibo.it}}
\runauthor{J. Mortera, P. Vicard and C. Vergari}
\affiliation{Universit\`a Roma Tre, Universit\`a Roma Tre and
Universit\`a di Bologna}
\address[A]{J. Mortera\\
P. Vicard\\
Dipartimento di Economia\\
Universit\`a Roma Tre\\
Via S. D'Amico 77\\
00145 Roma\\
Italy\\
\printead{e1}\\
\hphantom{E-mail: }\printead*{e2}} 
\address[B]{C. Vergari\\
Dipartimento di Scienze Economiche \\
Universit\`a di Bologna\\
Strada Maggiore 45 \\
40125 Bologna\\
Italy\\
\printead{e3}}
\end{aug}

\received{\smonth{11} \syear{2011}}
\revised{\smonth{9} \syear{2012}}

%
\begin{abstract}
We study an economic decision problem where the actors are two firms
and the Antitrust Authority whose main task is to monitor and prevent
firms' potential anti-competitive behaviour and its effect on the
market. The Antitrust Authority's decision process is modelled using a
Bayesian network where both the relational structure and the parameters
of the model are estimated from a data set provided by the Authority
itself. A number of economic variables that influence this decision
process are also included in the model. We analyse how monitoring by
the Antitrust Authority affects firms' strategies about cooperation.
Firms' strategies are modelled as a repeated prisoner's dilemma using
object-oriented Bayesian networks. We show how the integration of
firms' decision process and external market information can be modelled
in this way. Various decision scenarios and strategies are
illustrated.
\end{abstract}

%
\begin{keyword}
\kwd{Antitrust Authority}
\kwd{Bayesian networks}
\kwd{mergers}
\kwd{model integration}
\kwd{prisoner's dilemma}
\kwd{repeated games}
\end{keyword}

\end{frontmatter}

\section{Introduction}

Firms in many cases have incentives to cooperate (collude) to increase
their profits. The possibility for firms to collude does not depend solely
on their decision but also on external circumstances. First of all,
firms need to comply with antitrust laws. If the Antitrust Authority (AA)
finds negative anti-competitive effects, resulting from firms'
cooperative behaviour, it may intervene to prevent the firms from merging.

The AAs decision process is modelled here by using a Bayesian network
(BN) or Probabilistic Expert System (PES) [\citet{crgdaplssdj}]
estimated from real data. A BN is a graphical model that encodes the
probabilistic relationships among the variables of interest allowing
for the application of fast general-purpose algorithms to compute
inferences.

Often governments may find negative anti-competitive effects
resulting from a merger. As a consequence, the decision by firms
to cooperate is actually affected by the decision process of the
AA. The AA may start an investigation either because two firms
make a formal request to merge (explicit collusion) or because the
authority suspects that two firms are implicitly colluding. In
what follows the term merger will be used for both explicit and
implicit collusion.

We also study how the AAs monitoring affects firms' strategies about
cooperation. For this purpose, the firms' set of potential strategies are
modelled in turn as a repeated prisoner's dilemma using object-oriented
Bayesian networks (OOBNs) [\citet{kdpa,bowph}].
OOBNs are a recent extension of BNs which allow for a hierarchical
definition and construction of a BN. They provide a compact and
intuitive representation of the repeated prisoner's dilemma (PD).
Furthermore, thanks to the modularity and flexibility of this approach,
various sources of uncertainty within the game and generalizations of
the repeated prisoner's dilemma can be analysed. We use the PD as a
naive representation of firms' economic interaction, the focus of this
paper being that of analysing the evolution of firms' behaviour
according to various external scenarios. For theoretical aspects on
suboptimal strategies in Bayesian games see, for example,
\citet{youngsmith}.

We present two different networks: the first models the AAs
decision process, and the second represents the behaviour of the
two firms in a duopoly. OOBNs give the graphical framework to
integrate these two networks and to represent their time
evolution. Both the graphical structure and the associated
probability tables of AAs decision process network are estimated
from a real data set. As a result, we obtain the estimated
probability that AA intervenes to prevent anticompetitive
behaviour of a merger.
For various economic sectors (markets of interest) we study the
sensitivity of cooperative outcomes with
respect to factors such as geographical size, market share,
Herfindahl--Hirschman Index (HHI) variation, vertical effects, the
presence of entry
barriers and buyer power.
The global OOBN model which integrates the AAs decision process with a
duopoly model is used to obtain the optimal decision in light of a
series of interesting scenarios that could occur in practice.

The outline of the paper is as follows. In Section \ref{secproblem}
we briefly
describe the merger control problem. We illustrate the BN for the AAs
decision process estimated from the data and show its use in various
scenarios in Section \ref{secAAnet}. A brief introduction to the
prisoner's dilemma
is illustrated in Section \ref{secPD} followed by the Bayesian network
representation of the PD in Section \ref{secPDtoBN}. After
introducing the repeated
prisoner's dilemma in Section \ref{secRepeatedPD}, in Section \ref
{secOOBNPD} we show
how this can be represented as an OOBN. In Section \ref{secGlobal} we
show how we
integrate the PD network with the AA network obtaining a general
purpose global representation of the problem, and in Section \ref
{secfirms} we
apply this
to several decision scenarios. Finally, in Section \ref
{secconclusions} we
draw conclusions and discuss further developments.

\section{The merger control problem}
\label{secproblem}

The AA studies the impact of a merger on the market and its
consequences on social welfare. Hence, the AAs decision affects the
dynamics in
firms' economic interaction as well as the corresponding equilibrium
outcome. When choosing between cooperating or defecting, firms take the AAs
decision process into account, both when they formally request to merge
and in the case of implicit collusion.

\begin{figure}

\includegraphics{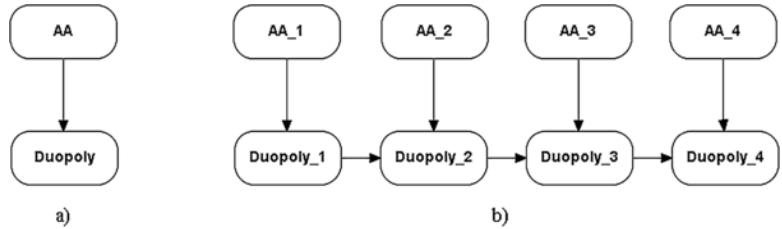}

\caption{\textup{(a)} Pictorial representation of the AA decision process and
Firms' behavior in a Duopoly. \textup{(b)} Corresponding representation for a
repeated scenario.}
\label{figintegrated}
\end{figure}

In our setup, the actors are as follows: the Antitrust Authority and
the two merging firms, termed Firm1 and Firm2 (the duopolists). Figure
\ref{figintegrated}(a) shows a pictorial representation of the effects
of AAs control activity on Firms' behaviour. The two rounded
rectangles, AA
and Duopoly, represent the AAs decision process and the Firms' merging
strategy, respectively. The AAs decision process is modelled by a Bayesian
network learned from real data (see Sections \ref{secAAnetwork} and
\ref{secAAprob}). The duopoly is modelled as a PD using a
Bayesian network for
decision making (see Section~\ref{secduopoly}). The two networks are then
integrated giving rise to a global model, where both
the AAs decision process and the duopoly are represented by OOBNs.
Figure \ref{figintegrated}(a) represents a single stage (vertical
slice) of the overall model. The merger problem, as well as AAs
activity, evolve in time. Figure \ref{figintegrated}(b) gives a graphical
representation of the decision process dynamics. Details on these
networks are given in Sections \ref{secduopoly} and \ref{secGlobal}.

\section{Antitrust Authority's decision process} \label{secAAnet}
\subsection{Current practice} \label{secCP}

The primary task of the AA is to enforce the antitrust law which
prohibits anticompetitive behaviour, so as to prevent a reduction in social
welfare.\setcounter{footnote}{1}\footnote{For details on the Italian antitrust law and AAs
tasks see: \url{http://www.agcm.it/en}.} In particular, the AA is
responsible for
detecting the following: (a)~agreements restricting competition; (b)
abuses of dominant positions; (c) merger operations involving the
creation or strengthening of
dominant positions in ways that eliminate or substantially reduce competition.

Once the Authority has received a complaint or has collected
information on possible interference with competition, a preliminary
examination is
carried out and if there are alleged violations of the Antitrust law,
the AA carries out a full investigation. The law requires that whenever the
potentially merging firms exhibit sale revenues in excess of certain
predefined thresholds, the merger operation must be notified to the authority
in advance. The thresholds are updated annually according to the
deflator index for gross domestic product.

\begin{table}
\caption{Description of the variables in the AA network}
\label{tabvariables}
\begin{tabular*}{\tablewidth}{@{\extracolsep{\fill}}lll@{}}
\hline
\textbf{Variable} & \multicolumn{1}{c}{\textbf{States}} & \multicolumn{1}{c@{}}{\textbf{Description}} \\
\hline
Years & \{1991--1996, 1997--2000, & Reference periods \\
& 2001--2003\} & \\
ATECO & Mining, food \& beverage & Relevant market \\
& Manufacture, etc. (see Figure \ref{figAAmonitor}) & \\
Geo size & \{Sub-national, national, & Size of the relevant market \\
& supra-national\} & \\
Buyer power & \{Yes, No\} & Presence (Yes) of competitive \\
& & pressure on the merging parties \\
Entry barriers & \{Yes, No\} & Presence (Yes) of entry barriers \\
HHI variations & \{0, $(0,100)$, [100, 500), & Variation in market \\
& [500, 1000), $\geq$1000 \} & concentration index \\
Post market share & \{$<$20\%, [20\%--40\%], $>$40\%\} & Post-merger
market share \\
Vertical effects & \{Yes, No\} & Presence (Yes) of vertical\\
& & effects \\
AA intervention & \{0, 1\} & No (0)/Yes (1) \\
\hline
\end{tabular*}
\end{table}

Decisions on a merger are based on a case by case examination and, to
our knowledge, currently, no specific models are used. The law also
does not
give any specific thresholds for relevant variables, such as market
share or a market concentration index.

\subsubsection{The data}
The data we use were collected by the Italian Antitrust Authority and
concern all the cases examined from 1991 to 2003. This data set
consists of
6920 observations. Based on this data set, \citet{lnmaervcf}
developed a logit model to analyse the impact of different factors on
the Authority
decision. Following \citet{lnmaervcf}, we consider relevant markets
affected by the merger as elementary units of analysis. These markets are
denoted by the ISTAT (Italian National Institute of Statistics)
economic activity code ATECO.

Table \ref{tabvariables} describes the variables in the data set that
were used to estimate the AA network. The Herfindahl--Hirschman
Index, HHI, is defined as the sum of the squares of $n$ firms' market
share, $\sum_{i}^n {\alpha_i}^2$, where $\alpha_i$ denotes firm
$i$'s market share and $\sum_{i}^n \alpha_i=100$. Increase in the
HHI indicates a decrease in competition and an increase in market
power. Vertical effects refer to the anticompetitive effects that
a vertical merger could imply, that is, the possibility to raise
entry barriers by input foreclosure or by customer foreclosure.

The estimation (learning) process of a Bayesian network consists
of two phases: the graphical structure estimation and the
conditional probability table estimation. These will be
illustrated in turn.

\subsection{Estimation of the network's graphical structure} \label
{secAAnetwork} The graphical structure of the AA network representing
the AA
decision process is obtained by a combination of subject-matter
knowledge, provided by a domain expert, and the information in the
data.

The \textit{Necessary Path Condition} (NPC) algorithm [\citet{steck}]
implemented in \textsc{Hugin} is used to estimate the graphical
structure of the
network. The NPC is a constraint-based algorithm recursively testing
marginal and conditional association between categorical variables. The NPC
algorithm allows the user to choose the most suitable among
independence equivalent models. The NPC algorithm takes into account logical
constraints, such as presence/absence of a link or assignment/ban of a
specific direction between variables.

\begin{figure}

\includegraphics{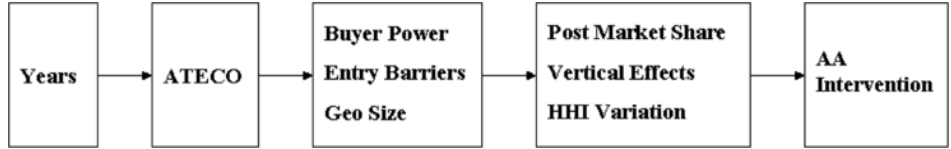}

\caption{Logical constraints for AA network estimation.}
\label{figlogical}
\end{figure}

The logical constraints we implemented here are shown in
Figure \ref{figlogical}. These imply that if there is a relation between
two variables in different boxes, it must have the same direction
as that in Figure \ref{figlogical}. Furthermore, if two variables belong
to the same box, their association (if it exists) can be\vspace*{1pt} in any
one of the two possible directions. For example, if node
\texttt{AA} \texttt{Intervention}\footnote{Here we indicate nodes in
\texttt{teletype}.} is connected with any of the other variables,
the direction has to be from these into \texttt{AA} \texttt{Intervention}
node (AA decision logically depends on the values of the other
variables). This means that arrows from \texttt{AA} \texttt{Intervention}
to any other variable are logically prohibited. The reference
period (node \texttt{Years}) is not influenced by any of the other
variables in the model.

\begin{figure}

\includegraphics{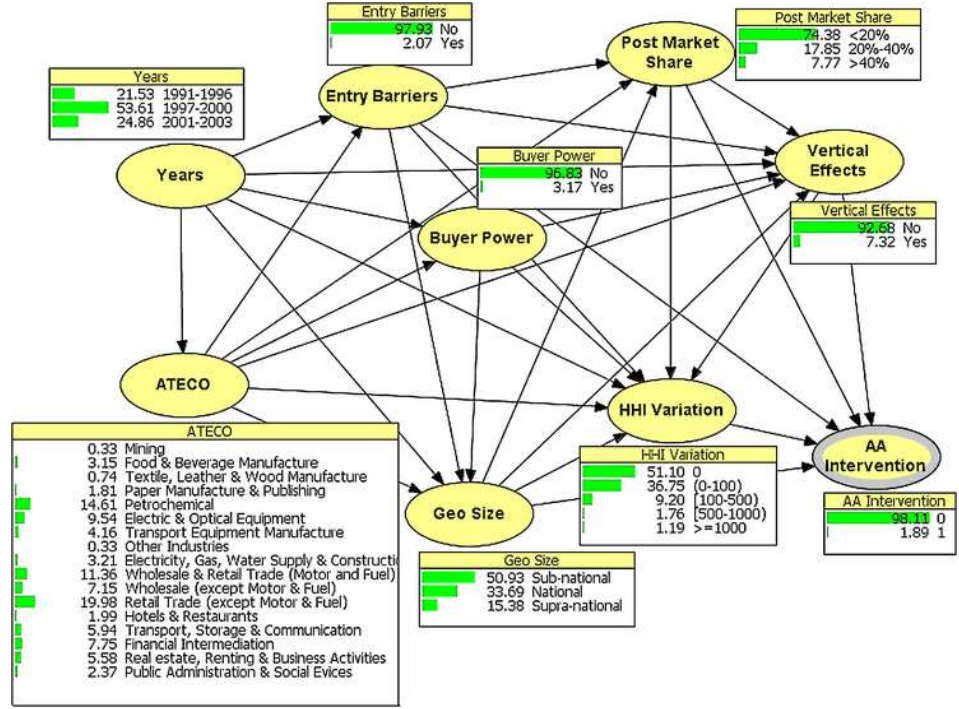}

\caption{AA network showing the dependencies of
\textit{\texttt{AA}} \textit{\texttt{Intervention}} on the relevant variables describing
the market and the marginal probabilities of the variables.}
\label{figAAmonitor}
\end{figure}

The dependence structure---based on the logical constraints
given in Figure \ref{figlogical}---learnt from the data is shown in
Figure \ref{figAAmonitor}. The main dependence relationships estimated
from the data are as follows:
\begin{longlist}[(iii)]
\item[(i)]The market of interest (\texttt{ATECO}) can depend on
\texttt{Year}: an economic sector could be more relevant and
worth investigating during one of the three reference periods (note that
the president of the AA changed in 1997 and from 2001 Italian
currency Lira was replaced by the Euro).
%
\item[(ii)] \texttt{AA} \texttt{Intervention} depends directly on
\texttt{HHI} \texttt{Variation}, \texttt{Vertical} \texttt{Effects},
\texttt{Post} \texttt{Market} \texttt{Share}, \texttt{Geo} \texttt{Size} and
\texttt{Entry} \texttt{Barriers}.
Furthermore, the relevant market (\texttt{ATECO}) does not affect AAs
decision (\texttt{AA} \texttt{Intervention}) directly but only through
the relevant features of the market and of the merging firms (\texttt
{HHI} \texttt{Variation}, \texttt{Vertical} \texttt{Effects},
\texttt{Post} \texttt{Market} \texttt{Share}, \texttt{Geo} \texttt{Size} and
\texttt{Entry} \texttt{Barriers}).

These results are consistent with those in \citet{bergman} and \citet
{lnmaervcf}.

\item[(iii)] The Herfindahl--Hirschman concentration index variation
(\texttt{HHI}\break \texttt{Variation}) depends on all the variables that
logically precede it or are on an equal footing (as shown in Figure
\ref{figlogical}), whereas \texttt{Post} \texttt{Market} \texttt{Share}
depends only on \texttt{Entry} \texttt{Barriers}, \texttt{Geo}
\texttt{Size} and \texttt{ATECO}. An explanation of this could be that
when a market sector is characterised by entry barriers (because of
patents or increasing returns to scale) we expect that this market may
be composed of a few firms with high market shares, thus influencing
\texttt{Post} \texttt{Market} \texttt{Share} and a relevant
\texttt{HHI} \texttt{Variation}.
\end{longlist}
Many other conditional independencies can be read off the AA
network in Figure~\ref{figAAmonitor}, but for brevity they will not be
presented here.

\subsection{Estimation of the probability tables} \label{secAAprob}
To complete the construction of our model, we estimate the conditional
probability distributions of the variables from the data. The
EM-algorithm [\citet{daplnmrdbjrss}] is used for learning the probabilities.

The network in Figure \ref{figAAmonitor} exhibits a complex association
structure among the variables. For example, node \texttt{HHI}
\texttt{variation} has seven parents. Its conditional probability table
has $17 \times3^3 \times2^3 \times5 = 18\mbox{,}360$ entries
corresponding to the state space of its parent variables:
\texttt{ATECO}, \texttt{Post} \texttt{Market} \texttt{Share},
\texttt{Years}, \texttt{Geo} \texttt{Size}, \texttt{Entry} \texttt{Barriers},
\texttt{Buyer} \texttt{Power}, \texttt{Vertical} \texttt{Effects}, as well as
\texttt{HHI} \texttt{Variation}'s state space. Many of these combinations are
not represented in the data set, although they cannot be considered
impossible \textit{ex ante}. In fact, according to
\citet{bergman}, if a threshold for relevant variables---like post
market share---can be detected in AAs legal practice, this threshold
may vary according to other variables, such as buyer power and entry
barriers. Therefore, no variable level combinations can in principle be
ruled out. So, in order to avoid that certain possible configurations
in the conditional probability tables have zero probability, we set
noninformative nonzero prior probabilities.

Figure \ref{figAAmonitor} displays the marginal
probabilities\footnote{In all figures probabilities are expressed
as percentages.} estimated from our data. Note, for example, that
the probability of an AA intervention is only $0.0189$, which could
be due to the fact that in most cases, $74.38\%$, the post market
share is less than $20\%$ and entry barriers and vertical effects
are absent (with probability $0.9793$ and $0.9268$, resp.),
HHI index is less than 100 in $87.85\%$ of the cases and only in
$15.38\%$ the geographical size is supra-national.
%

\subsection{Using the network} Once the model has been estimated, we
can address a number of questions about the AAs decision process. Various
possible scenarios can be examined by inserting and propagating the
appropriate evidence throughout the network.
We illustrate three hypothetical scenarios.

\begin{figure}

\includegraphics{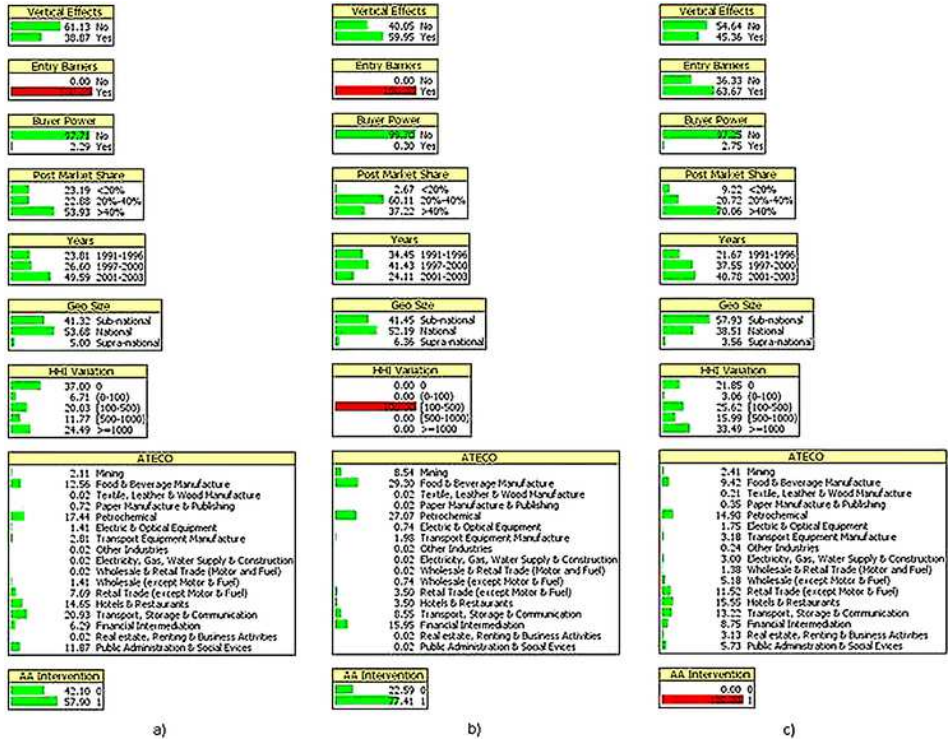}

\caption{Scenarios \textup{(a)}, \textup{(b)} and \textup{(c)} giving marginal posterior
probabilities for the AA network.} \label{figPostAI}
\end{figure}

\textit{Scenario} A. What is the probability of an AA intervention in a
merger request when there are entry barriers in the market? This
scenario is represented in Figure~\ref{figPostAI}(a). The posterior
probability of an \texttt{AA} \texttt{Intervention} increases from $0.0189$ to
$0.5790$ when the evidence \texttt{Entry} \texttt{Barriers}${}={}$Yes is
inserted and propagated throughout the network.

%

\textit{Scenario} B. How would the probability obtained in
Scenario A change if the Herfindahl--Hirschman concentration
index\vadjust{\goodbreak}
variation (\texttt{HHI} \texttt{variation}) is in the class $[100, 500)$?
Note in Figure \ref{figPostAI}(b) that the probability of
\texttt{AA} \texttt{Intervention} now increases to $0.7741$.

The network can be used not only for direct reasoning about the
probability of \texttt{AA} \texttt{Intervention}, but also for reasoning about possible
``causes'' of a given AA decision.

\textit{Scenario} C. A question about competition authorities'
behaviour that has been rarely addressed in the literature is
about the type of mergers that are typically prohibited
[\citet{bergman}]. Our network can be used for this purpose. Suppose
that the AA decides to intervene in a firm's merger request. What
are the most plausible reasons of this decision? Figure \ref{figPostAI}(c)
gives the posterior probabilities given the evidence that
\texttt{AA} \texttt{Intervention} is equal to one. On comparing
Figures \ref{figAAmonitor} and \ref{figPostAI}(c) we see that:
\begin{itemize}
\item The probability of entry barriers increases from 0.0207 to 0.6367;
\item The probability of vertical effects increases from 0.0732 to 0.4536.
This is an interesting result, since, although there is common
agreement about the relevance of vertical
effects for AAs decision on a merger request, it is controversial
whether vertical effects influence
the market negatively by foreclosing competitors or positively by
reducing transaction costs.
Here we find that the presence of vertical effects is much more
probable for those firms where AA decides to
intervene. \citet{lnmaervcf} found similar results.
\item The probability of post market share less than $20\%$ decreases
from 0.7438 to 0.0922, whereas the
probability of post market share greater
than $40\%$ increases from 0.0777 to 0.7006.
\item The HHI index decreases in the first two classes and increases in
the last three classes.
\end{itemize}
Note that when evidence is propagated in the network, all marginal
probability tables are updated accordingly.

%
\begin{table}
\tablewidth=125pt
\caption{Payoff matrix for the prisoner's dilemma}
\label{tabNormalForm1}
\begin{tabular*}{\tablewidth}{@{\extracolsep{\fill}}lccc@{}}
\hline
& & \multicolumn{2}{c}{Firm2} \\
& & $C$ & $D$ \\
\multirow{2}{*}{Firm1} & $C$ & $a,a$ & $c,d$ \\
& $D$ & $d,c$ & $b,b$\\
\hline
\end{tabular*}
%
\end{table}
%

\section{Duopoly representation}
\label{secduopoly}

\subsection{The prisoner's dilemma} \label{secPD}
The prisoner's dilemma [\citet{floodms}]
describes cooperation by rational agents. The PD is a 2-player
symmetric game where the two players have the same r\^{o}le and have
the same
set of potential strategies termed \textit{cooperate} $C$ and
\textit{defect} $D$. The PD is a simultaneous game where the players choose
just once
and simultaneously and the unique equilibrium\footnote{An
equilibrium is a strategy pair such that no player can improve his
position by
unilaterally changing his decision. In other words, it is a situation
in which all players choose mutual best responses.} is the pair of
strategies ($D$, $D$). Players' payoffs are such that defect is a
dominant strategy, that is, a strategy that is preferred by each player
independently
of his/her rival. The problem is that this strategy is inefficient
since both players would gain more if they cooperated and adopted the ($C$,
$C$) strategy. The source of the dilemma lies in the fact that each
player has an incentive to defect if the rival player cooperates, so
that an
agreement to cooperate would not be credible.

Simultaneous games, such as the PD, are commonly represented in
either the normal or the extensive form. In the normal form
representation, the PD can be described by the payoff matrix in
Table \ref{tabNormalForm1}. The two firms, Firm1 and Firm2,\vadjust{\goodbreak} have two
available strategies: cooperate $C$ or defect $D$. The payoffs
need to be such that $d>a>b\geq c$ and $2a>(c+d)>2b$, so that
$(C,C)$ maximises players' joint payoff. Given that $b<a$, the
strategy pair $(D,D)$ is strictly worse than $(C,C)$.

\begin{figure}

\includegraphics{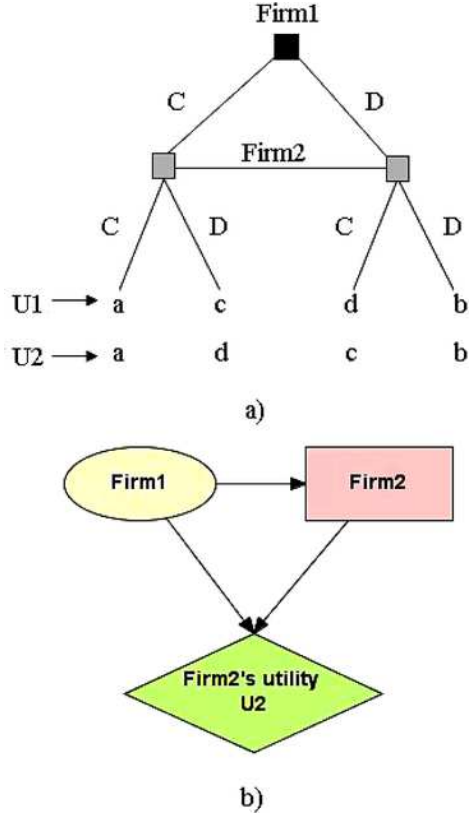}

\caption{\textup{(a)} Tree representation of the simultaneous duopoly game. \textup{(b)}
Corresponding Bayesian network representation.}
\label{figgame}
\end{figure}

In the extensive form the game is represented by a tree.
Figure \ref{figgame}(a) shows the tree representation (equivalent to
Table \ref{tabNormalForm1}) of the simultaneous duopoly game. Firm1
moves first and chooses either $C$ or $D$, Firm2 moves second but
without knowing what Firm1 did.

A symmetric duopoly, such as a market with two symmetric
profit-maxi\-mising firms in mutual competition, can be modelled
as a PD. The duopoly profit is the gain of each of the sellers in
this market.

Suppose the two firms produce identical goods, incurring constant
margi\-nal costs, and they compete setting their prices. Since
consumers will buy from the firm charging the lowest price, firms
have an incentive to undercut their price to conquer the market
(noncooperative or defect strategy). At equilibrium firms will
set the competitive price (the market price under perfect
competition which is equal to firm's marginal cost of production),
gaining duopoly profit $b=0$.
This result is often called a paradox, since there are just two firms
in the market and still the perfectly competitive strategy yields zero
profit. However, if firms decide to cooperate and set the monopoly
price, they can share positive monopoly profits. The monopoly profit is always
greater than twice the duopoly profit, $2a>2b$.

In most markets, from a consumer's point of view, goods are not
identical. This gives firms the ability to raise the price above the
marginal cost
of production without losing their customers to competitors. In a
symmetric duopoly with product differentiation firms produce and sell
differentiated goods (imperfect substitutes). As long as product
differentiation is not too large, firms face a PD: if they cooperate,
they could
share monopoly profit, but they have incentive to defect if the rival
cooperates. However, when goods are imperfect substitutes, firms make
positive duopoly profit, $b>0$, under the noncooperative strategy pair
$(D,D)$. This duopoly profit is smaller than half the monopoly profit,
$b<a$, so that the cooperative strategy $C$ is superior for each firm singly.

\subsection{The prisoner's dilemma network}
\label{secPDtoBN}

Bayesian networks for decision support systems can incorporate both
decision nodes and utility nodes [\citet{jensen}], giving rise to an influence
diagram (ID) representation. IDs were extended by \citet{lslndms} to
allow for limited information decision problems (LIMIDs). A different
approach to represent and solve games using graphical models was
initially proposed by \citet{smith} and later by \citet{lamura},
\citet{kmlmss} and \citet{kdmbgeb}.

The one stage PD being a symmetric game can be represented by the
ID network in Figure \ref{figgame}(b). The simultaneity of the game is
implemented by representing Firm1 as a random variable (oval
node) and Firm2 as the decision maker (rectangular node) having
two possible actions:
defect $D$ and cooperate $C$. Firm2's decision is influenced by
Firm1. Firm1's associated prior probability distribution
represents Firm2's subjective opinion about Firm1's behaviour.
Random variable Firm1 has two states, defect (coded as 0) and
cooperate (coded as 1), with uniform prior probabilities
indicating Firm2's ignorance about Firm1's choice. Firm2 could
assign different prior probabilities based on his/her prior
knowledge about Firm1's behaviour. Table \ref{tabutilitytable} shows
Firms2's utility [node \texttt{Firm2's} \texttt{utility} \texttt{U2} in
Figure \ref{figgame}(b)] based on Firm1 and Firm2's actions. Thanks to game
symmetry, Table \ref{tabutilitytable} is equivalent to the normal form
payoff matrix given in Table \ref{tabNormalForm1}.

\begin{table}
\tablewidth=275pt
\caption{Firm2's utility \textit{\texttt{U2}} conditional on \textit{\texttt{Firm1}} and
\textit{\texttt{Firm2}}'s actions}
\label{tabutilitytable}
\begin{tabular*}{\tablewidth}{@{\extracolsep{\fill}}lcccc@{}}
\hline
\textbf{\texttt{Firm1}} & \multicolumn{2}{c}{\textbf{Defect (0)}} & \multicolumn
{2}{c@{}}{\textbf{Cooperate (1)}} \\[-4pt]
& \multicolumn{2}{c}{\hrulefill} & \multicolumn{2}{c@{}}{\hrulefill}\\
\textbf{\texttt{Firm2}}& \textbf{Defect (0)} & \textbf{Cooperate (1)}
& \textbf{Defect (0)} & \textbf{Cooperate (1)} \\
\hline
\texttt{U2} & $b$ & $c$ & $d$ & $a$ \\
\hline
\end{tabular*}
\end{table}

Once the network is compiled, the optimal decision for Firm2 is
automatically computed by maximising expected utility. Since the
game is symmetric, Firm2's optimal strategy coincides with
Firm1's optimal strategy and this pair of strategies constitutes a
Nash equilibrium. Thus, in the ID representation the choice of
Firm2 as decision maker is without loss of generality.\eject

In what follows we always consider Firm2 as the decision maker.
The prior probability distribution on the random variable Firm1
reflects Firm2's subjective opinion on the type of rival player
he/she is playing against.


\subsection{Repeated prisoner's dilemma} \label{secRepeatedPD}

Since firms interact more than once, we need to consider the repeated
version of the PD. In repeated games, players' actions are observed at
the end
of each period and their overall payoff is the sum of the payoffs in
each stage discounted by a factor $\delta\in[0,1]$. Thus, players may
condition their play on the opponents past play. Here we assume that
firms never forget previous moves and other information acquired, in other
words, we assume that firms have perfect recall.

The repeated PD analyzes how threats and promises about future
behaviour can affect and improve current behaviour. When the time
horizon is
indefinite firms may decide to adopt a cooperative strategy where the
discount factor $\delta$ represents uncertainty about
the number of stages faced by firms. This uncertainty is usually not
modelled within the game itself. In Section \ref{secGlobal} we
illustrate how to
incorporate this uncertainty in the merger control problem.

\subsection{OOBN for repeated prisoner's dilemma}
\label{secOOBNPD} 
\begin{figure}

\includegraphics{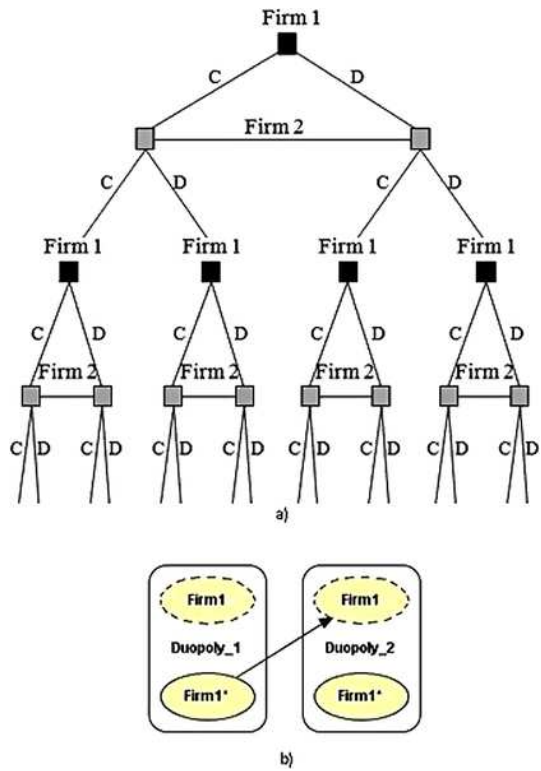}

\caption{\textup{(a)} Tree representation of the two-stage duopoly game. \textup{(b)}
Corresponding OOBN representation.}
\label{figalberoPD}
\end{figure}

Generalising the tree representation in Figure
\ref{figgame}(a) to repeated games is both computationally and
graphically demanding. The game tree grows exponentially with the
number of stages. For example, Figure \ref{figalberoPD}(a) shows the
tree representation of a two-stage PD.

OOBNs are particularly well suited for an application area such as the
present because the
similarity between network elements (the stages of the game) can be
exploited in a modular and flexible construction. Object-oriented Bayesian
networks have a hierarchical structure where a node itself can
represent a (object-oriented) network containing several
\textit{instances} of other
generic \textit{classes} of networks. Instances have interface
\textit{input} and \textit{output} nodes as well as ordinary nodes. Instances
of a
particular class have identical conditional probability tables for
noninput nodes. Instances are connected by arrows from output nodes
into input
nodes. These arrows, as well as those from ordinary nodes to input
nodes, represent identity links, whereas arrows between two ordinary
nodes or
an output node and an ordinary node represent\vadjust{\goodbreak} probabilistic dependence.
The graphical simplicity automatically
produces computational efficiency. As a result, increasingly
complex networks can be constructed by simply adding new objects
which perform different tasks.

\begin{figure}

\includegraphics{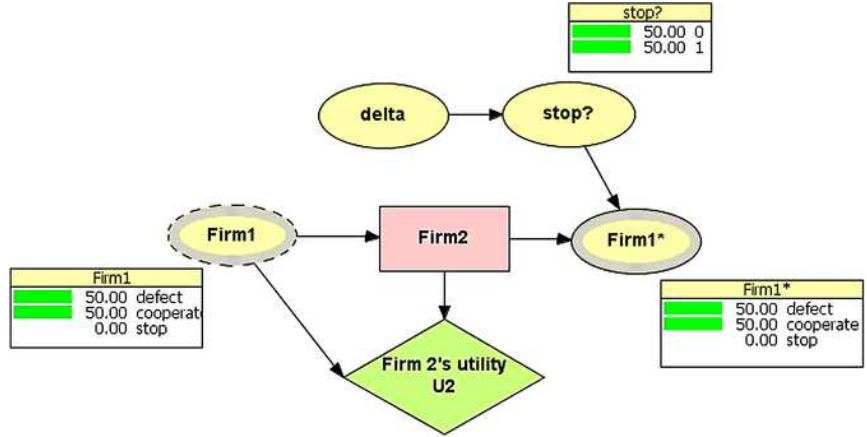}

\caption{Class network for repeated PD with associated marginal prior
probability tables.}
\label{figModifiedPD}
\end{figure}

Since we assume perfect recall,
\textsc{Hugin}\footnote{\href{http://www.hugin.com}{www.hugin.com}.} version 6.9 software,
which automatically implements the fact that at every stage the
decision maker recalls all previous decisions, is used to build
the networks. This implies that each decision depends on the
decisions taken in all previous stages, so even though the
graphical representation does not implicitly represent this, in
the junction tree construction [\citet{crgdaplssdj}] these
dependencies are explicitly considered. In what follows we
indicate an instance in \textbf{bold}. Figure \ref{figalberoPD}(b) shows
the OOBN two-stage repeated game that corresponds to the tree
representation in Figure \ref{figalberoPD}(a). Each rounded rectangle
represents an instance termed \textbf{Duopoly} and models a stage
of the repeated game. In order to specify the links between
successive stages (instances), Figure \ref{figgame}(b) (which represents
each \textbf{Duopoly} instance) needs to be generalised as shown
in Figure \ref{figModifiedPD}.

%

\begin{table}[b]
\tablewidth=290pt
\caption{Conditional probability table for \textit{\texttt{Firm1$^*$}} given
\textit{\texttt{stop?}} and \textit{\texttt{Firm2}}}
\label{tabprobabilitytable}
\begin{tabular*}{\tablewidth}{@{\extracolsep{\fill}}lcccc@{}}
\hline
\textbf{\texttt{Stop?}} & \multicolumn{2}{c}{\textbf{No (0)}}
& \multicolumn{2}{c}{\textbf{Yes (1)}} \\
[-4pt]
& \multicolumn{2}{c}{\hrulefill} & \multicolumn{2}{c@{}}{\hrulefill}\\
\textbf{\texttt{Firm2}}& \textbf{Defect (0)} & \textbf{Cooperate (1)}
& \textbf{Defect (0)} & \textbf{Cooperate (1)}\\
\hline
Defect (0) & 1 & 0 & 0 & 0 \\
Cooperate (1) & 0 & 1 & 0 & 0 \\
Stop (2) & 0 & 0 & 1 & 1 \\
\hline
\end{tabular*}
\end{table}

The node \texttt{Firm1$^*$} models the behaviour of Firm1 in the next
stage. In each stage the game can either continue or terminate.
\texttt{Firm1} and \texttt{Firm1$^*$} now need to be given three
states: defect (0), cooperate (1) and stop (2). Since in a repeated
game every stage depends on the actions taken in the previous stages,
\texttt{Firm1$^*$} is logically dependent on \texttt{Firm2}.
Uncertainty about the existence of further stages is modelled by adding
a new random node \texttt{stop?}. Node \texttt{stop?} has two states, $
\{ 0,1 \}$ according to whether the game continues or stops and has a
Bernoulli distribution $\operatorname{Bin}(1,1-\mbox{\texttt{delta}})$.
The parameter node \texttt{delta} is the probability that the game
continues \mbox{$P(\mbox{\texttt{stop?}}=0)$}. Node \texttt{delta} has a uniform
prior distribution over a plausible set of values.

In the first stage, to ensure that the game starts, Firm1 can only
choose between defect and cooperate. Table \ref{tabprobabilitytable}
gives the conditional probability distribution of
\texttt{Firm1$^*$} given \texttt{stop?} and \texttt{Firm2}. It
shows that if the game stops (\texttt{stop?}${}={}$1), \texttt{Firm1$^*$}
stops with certainty, else \texttt{Firm1$^*$} cooperates or
defects according to \texttt{Firm2}'s decision. This implements
the \textit{tit for tat} (TFT) strategy in which Firm1 begins by
cooperating and cooperates as long as Firm2 cooperates, and
defects otherwise. Variations on this strategy will be shown in
Section \ref{secOS}.




\subsubsection{Other strategies}
\label{secOS}

\begin{figure}

\includegraphics{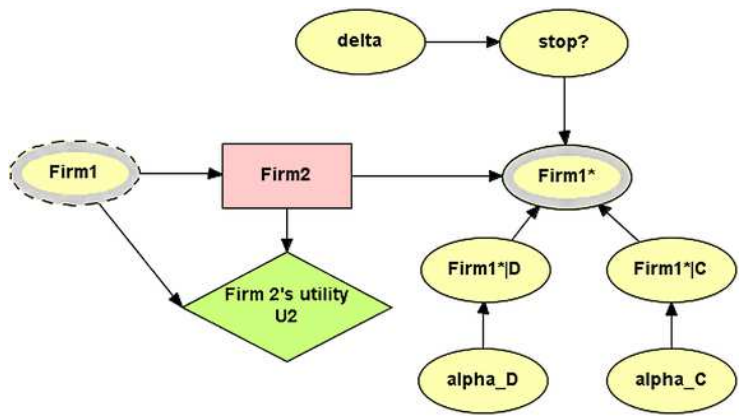}

\caption{Generalised repeated PD network representing various
strategies and incomplete information.}
\label{figfinalPD}
\end{figure}

Experimental results show that people, contrary to standard
prescriptions of game theory, may cooperate more frequently than expected
[\citet{ajmjhej}]. An explanation behind this empirical evidence is
provided by the theoretical models of \citet{kdmwrjet} and
\citet{kdmmprwrjet}.
Figure \ref{figModifiedPD} can be modified to provide a general class network
that explicitly incorporates a set of potential strategies for
Firm1 other than TFT. This network is displayed in
Figure \ref{figfinalPD}. The network can, for example, model a repeated
PD with incomplete information, that is, where there is
uncertainty about the type of rival that a firm is going to face.
The conditional probability distribution of Firm1$^*$ reflects Firm2's
uncertainty about its opponent. If Firm2 believes Firm1
to be ``altruistic'', it can expect Firm1 to cooperate, with
probability $\alpha_D>0$, even if it defected in the previous stage.
On the other
hand, if Firm2 believes Firm1 to be ``egoistic'', then it expects Firm1
to cooperate, with probability $\alpha_C <1$, even if it cooperated in the
previous stage.

Additional nodes, \texttt{Firm1$^*$|D} and \texttt{Firm1$^*$|C},
having Bernoulli distributions with parameter nodes
\texttt{alpha\_D} and \texttt{alpha\_C} are added to the network
of Figure~\ref{figModifiedPD}. Node \texttt{Firm1$^*$} takes value 2 if
the game stops in the current stage, whereas if the game
continues (\texttt{stop?}${}={}$0), the value of \texttt{Firm1$^*$}
depends on that of \texttt{Firm2}. If \texttt{Firm2} defects
(cooperates), \texttt{Firm1$^*$} is \texttt{Firm1$^*$|D}
(\texttt{Firm1$^*$|C}), with \texttt{alpha\_D} (\texttt{alpha\_C})
being the probability that Firm1 will cooperate in the next stage
given that Firm2 defected (cooperated) in the previous stage. The
conditional probability distribution of \texttt{Firm1$^*$} is thus
defined\vspace*{1pt} by the logical expression $\mbox{\textit{if}}(\mbox{\texttt{stop}}
== 1, 2, \mbox{\textit{if}}(\mbox{\texttt{Firm2}} == 0,
\mbox{\texttt{Firm1$^*$|D}},\break \mbox{\texttt{Firm1$^*$|C}}))$.\footnote{The
function $\mbox{\textit{if}}(A,x,y)$ takes value $x$ if condition
$A$ is satisfied, otherwise $y$.} Firm1$^*$ represents Firm2's
subjective opinions about Firm1's behaviour in each single stage
of the repeated game.



This model can also incorporate a large set of strategies, including
TFT, and it can model scenarios where the probability that the game
continues depends on external factors. An illustrative example is given
in Section \ref{secGlobal}.

\begin{figure}

\includegraphics{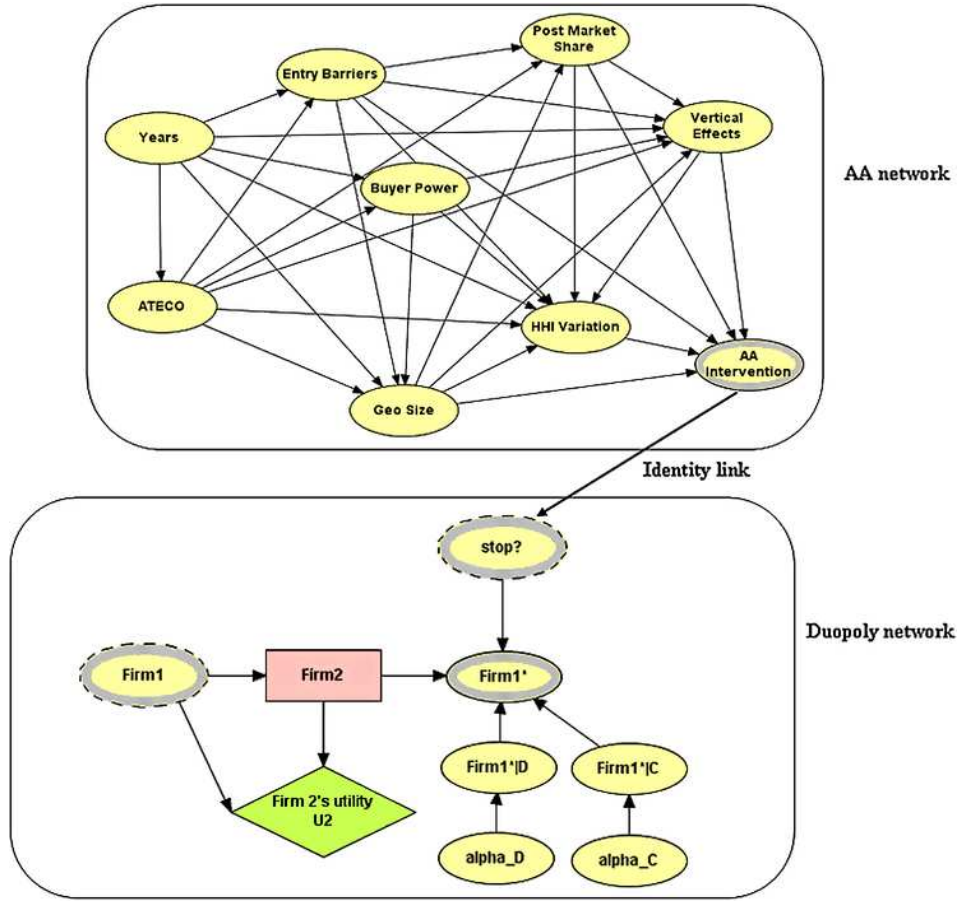}

\caption{Integrated AA-duopoly merger stage game.}
\label{figglobal}
\end{figure}

\section{Global network} \label{secGlobal} Thanks to the modularity
and flexibility of OOBNs, it is possible to integrate the AA and
the Duopoly networks, giving rise to a unique overall OOBN
representation of the problem [Figure \ref{figintegrated}(a)]. An expanded
representation of this model is shown in Figure \ref{figglobal}.\eject

The \textbf{Duopoly} network (the bottom network) in Figure \ref
{figglobal} is similar to the network in Figure \ref{figfinalPD} except
that the uncertainty about the next stage \texttt{stop?} is now
identified with \texttt{AA} \texttt{Intervention} in the \textbf{AA}
network (the top network in Figure~\ref{figglobal}) representing AAs
decision process.

The AA decision process is usually dynamic; it can change over
time due to changes in the antitrust law as well as changes in
market conditions. We are thus interested in the repeated version
of the model in Figure \ref{figglobal}.

\begin{figure}

\includegraphics{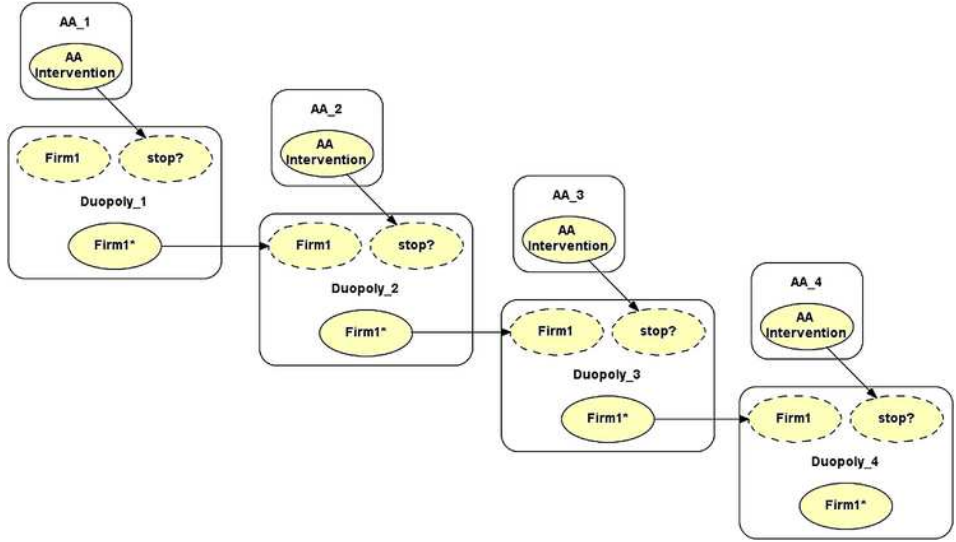}

\caption{OOBN representing a three-stage repeated merger game with
uncertainty about the number of stages.}
\label{fig3stages}
\end{figure}

Figure \ref{fig3stages} represents the global model (Figure \ref{figglobal})
repeated four times for a three-stage merger game with uncertainty
on the number of stages. In general, an OOBN with $n+1$ instances
models a game repeated $n$ times with uncertainty about the
successive stage. In this model, the AAs decision process is
represented by the same instance in each period. This is justified
by assuming that, even if the AA decides not to intervene, it
continues monitoring firms' behaviour in successive stages.

\subsection{Firms' strategy}\label{secfirms}


We now study the sensitivity of cooperative behaviour with respect
to two sets of utilities and all the factors that might directly
or indirectly influence the AAs decision. We consider both the
TFT strategy and a more general strategy. The TFT strategy can be
implemented using the global network by setting
\texttt{Firm1}$^*=1$ in stage \textbf{Duopoly\_1} and
\texttt{Firm1$^*$|C}${}={}$1, \texttt{Firm1$^*$|D}${}={}$0 in all other
stages.\newpage

\subsubsection{TFT strategy: Perfect substitutability}\label{secTFTperfect}
\begin{figure}[b]

\includegraphics{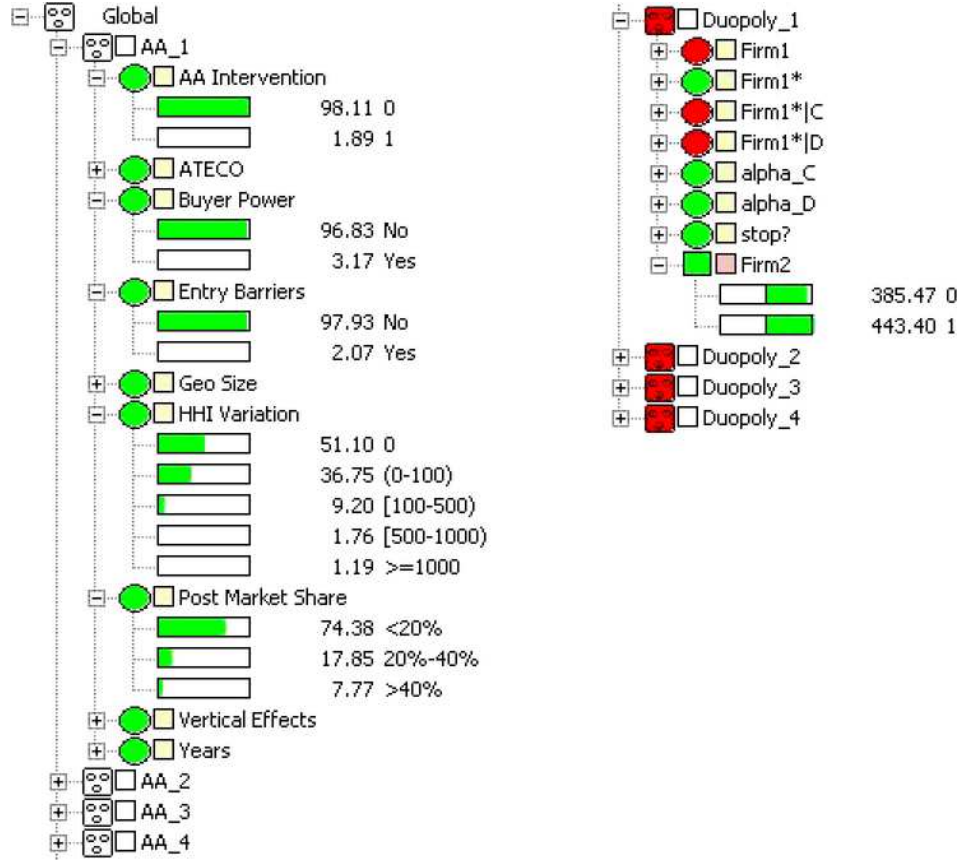}

\caption{Marginal probabilities and optimal decision in the first
stage \textbf{AA\_1} and \textbf{Duopoly\_1}, under perfect
substitutability, when Firm1 plays TFT.}
\label{figTFT}
\end{figure}

\begin{figure}

\includegraphics{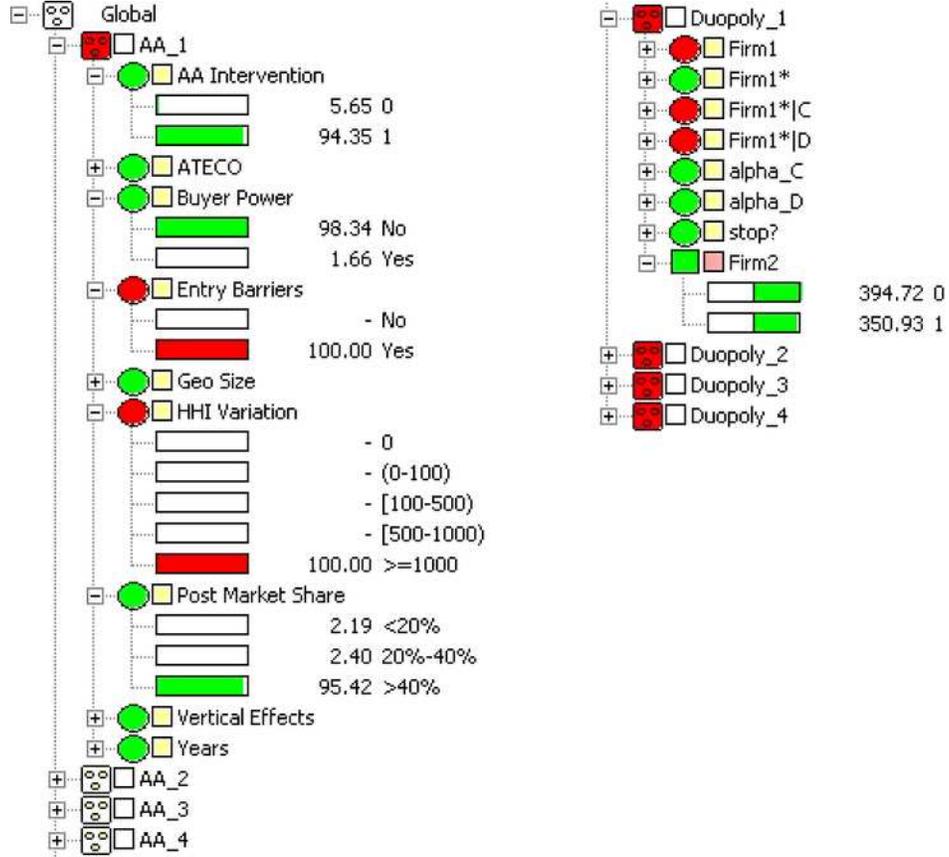}

\caption{Marginal probabilities and optimal decision
in the first stage \textbf{AA\_1} and \textbf{Duopoly\_1}, under
perfect substitutability, when Firm1 plays TFT, \textit{\texttt{Entry}}
\textit{\texttt{Barriers}}${}={}$Yes and \textit{\texttt{HHI}}
\textit{\texttt{Variation}}${}>={}$1000.} \label{figTFT1}
\end{figure}

\begin{figure}

\includegraphics{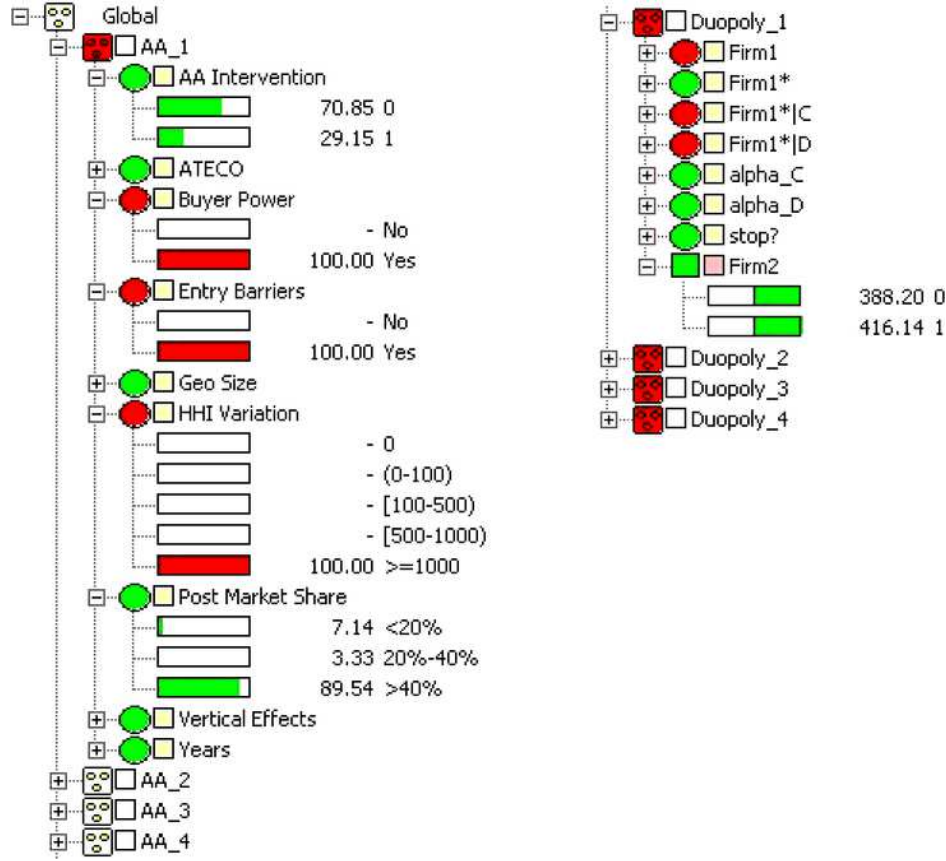}

\caption{Marginal probabilities and optimal decision
in the first stage \textbf{AA\_1} and \textbf{Duopoly\_1}, under
perfect substitutability, when Firm1 plays TFT, \textit{\texttt{Entry}}
\textit{\texttt{Barriers}}${}={}$Yes, \textit{\texttt{HHI}}
\textit{\texttt{Variation}}${}>={}$1000 and
\textit{\texttt{Buyer}} \textit{\texttt{Power}}${}={}$Yes.} \label{figTFT2}
\end{figure}

Table \ref{tabps} shows an example of Firm2's utility for a market with
perfect substitutable goods.
Figures \ref{figTFT}, \ref{figTFT1} and \ref{figTFT2} show the marginal
probabilities for a selection of random variables and the expected
utilities for the decision nodes in the first stage \textbf{AA\_1}
and \textbf{Duopoly\_1}.

%
\begin{table}
\tablewidth=270pt
\caption{Firm2's utility \textit{\texttt{U2}} for a market with perfect substitutability}
\label{tabps}
\begin{tabular*}{\tablewidth}{@{\extracolsep{\fill}}lcccc@{}}
\hline
\textbf{\texttt{Firm1}} & \multicolumn{2}{c}{\textbf{Defect (0})} & \multicolumn
{2}{c@{}}{\textbf{Cooperate (1)}} \\[-4pt]
& \multicolumn{2}{c}{\hrulefill} & \multicolumn{2}{c@{}}{\hrulefill}\\
\textbf{\texttt{Firm2}}& \textbf{Defect (0)} & \textbf{Cooperate (1)}
& \textbf{Defect (0)} & \textbf{Cooperate (1)} \\
\hline
\texttt{U2} & 0 & $-$10 & 150 & 100 \\
\hline
\end{tabular*}
\end{table}
%

When no evidence about the variables in the market is inserted in
the network (Figure \ref{figTFT}) Firm2's optimal decision is to
cooperate (1), having expected utility equal to 443.40 (while
defect has expected utility equal to 385.47). This could be in
part due to the small probability of an AA intervention, 0.0189.

Figure \ref{figTFT1} shows the case where there are entry barriers in
the market of interest (\texttt{Entry} \texttt{Barriers}${}={}$Yes) and
the merger causes the HHI variation to be in the last class
(\texttt{HHI} \texttt{Variation}${}>={}$1000). The resulting
probability of AA intervention shoots up to 0.9435 and Firm2's optimal
decision is to defect with expected utility of 394.72, against 350.93
for cooperating. This strategy still remains optimal (although with a
smaller gap between the expected utilities) when based only on the
presence of entry barriers.

Figure \ref{figTFT2} shows the case where, as before, there are entry
barriers, the HHI variation is $\geq$1000, and customers exert
competitive pressure on the merging parties (\texttt{Buyer}
\texttt{Power}${}={}$Yes). The probability of AA intervention decreases
from 0.9435 to 0.2915 and Firm2's optimal decision is to cooperate,
having expected utility of 416.14. It is interesting to note that buyer
power is able to counterbalance the effect of both entry barriers and a
large HHI variation.

\subsubsection{TFT strategy: Imperfect substitutability}
We now use Firm2's utility for a market with imperfect
substitutable goods given in Table \ref{tabps1}.

\begin{figure}[b]

\includegraphics{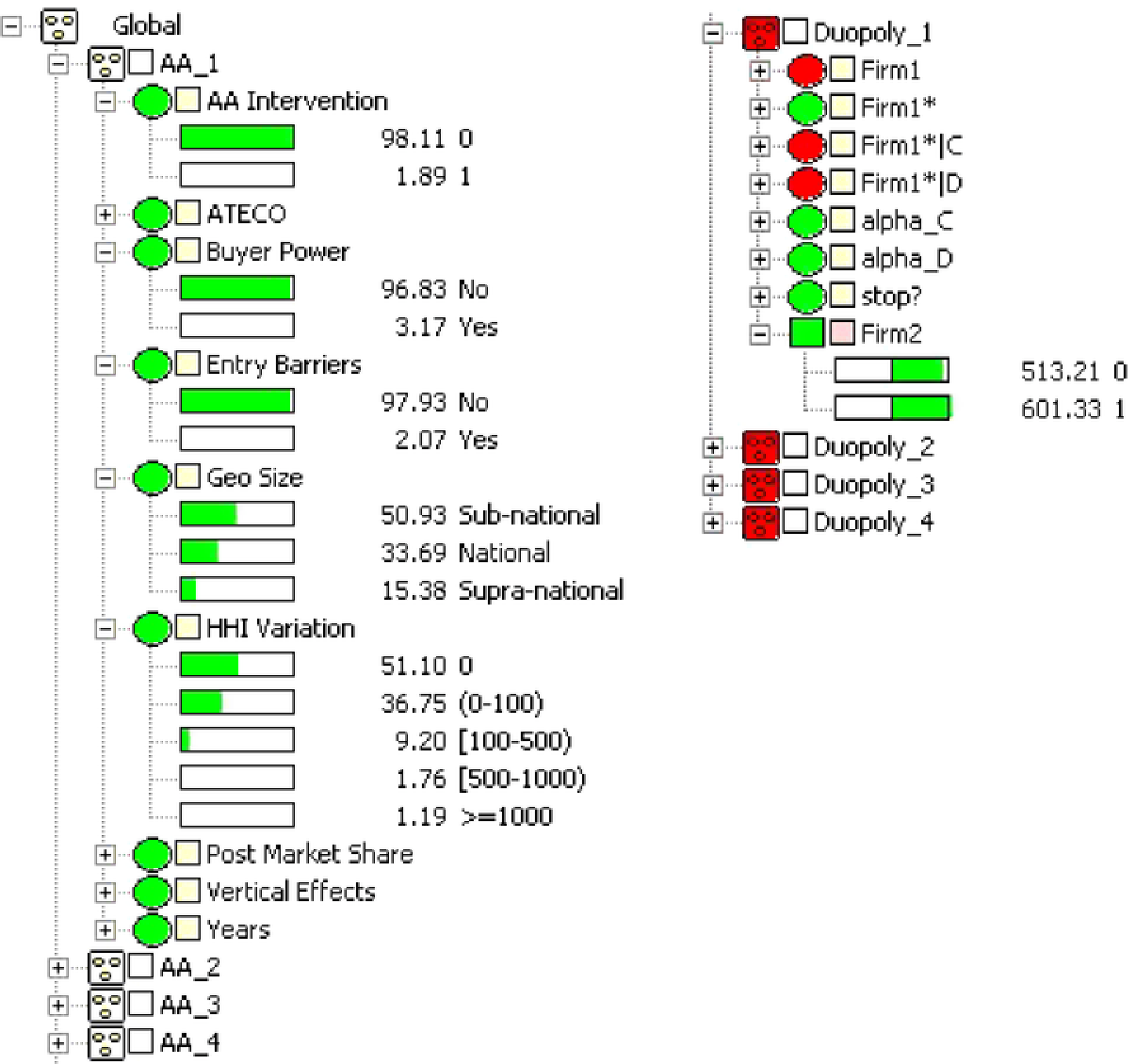}

\caption{Marginal probabilities and optimal decision in the first
stage \textbf{AA\_1} and \textbf{Duopoly\_1}, under
imperfect substitutability, when Firm1 plays TFT.}
\label{figTFTnoPS}
\end{figure}

Figure \ref{figTFTnoPS} shows results when evidence about the market is
not available. \texttt{Firm2}'s optimal decision is to cooperate
(1), having expected utility equal to 601.33 (while defect has
expected utility equal to 513.21). Again, this is most plausibly
due to the small probability of an AA intervention.

\begin{figure}

\includegraphics{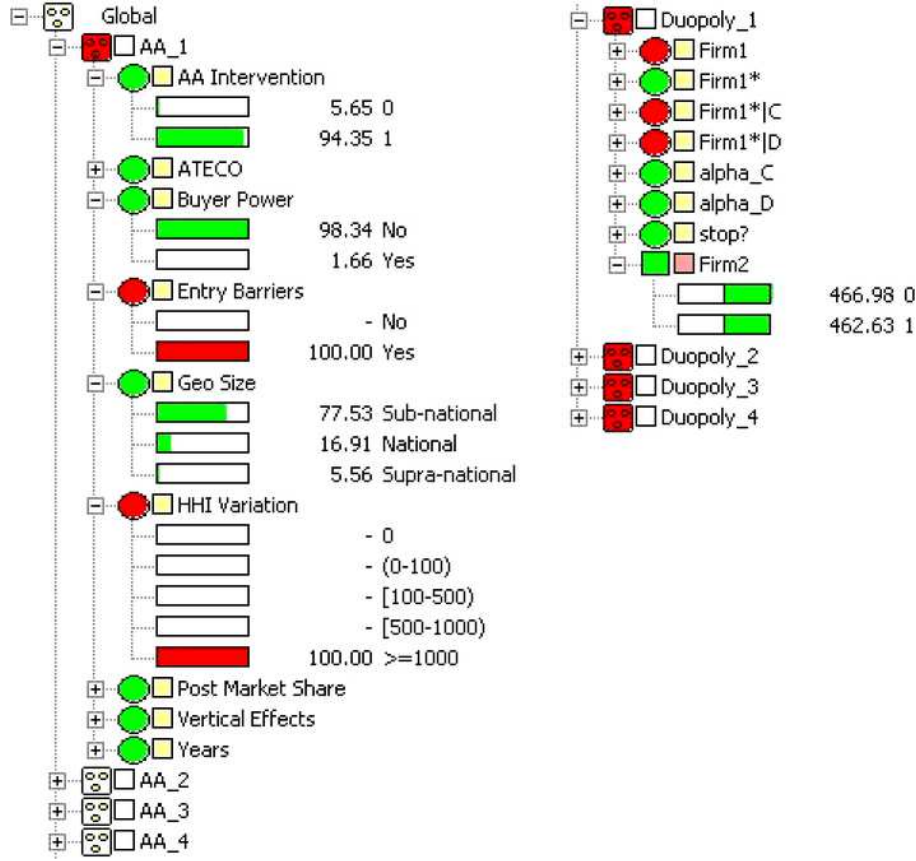}

\caption{Marginal probabilities and optimal decision
in the first stage \textbf{AA\_1} and \textbf{Duopoly\_1}, under
imperfect substitutability, when Firm1 plays TFT,
\textit{\texttt{Entry}}
\textit{\texttt{Barriers}}${}={}$Yes and \textit{\texttt{HHI}}
\textit{\texttt{Variation}}${}>={}$1000.} \label{figTFTnoPS1}\vspace*{-5pt}
\end{figure}

When \texttt{Entry} \texttt{Barriers}${}={}$Yes and \texttt{HHI}
\texttt{Variation}${}>={}$1000, Firm2's expected utility to cooperate
or to defect is almost equal, although the probability of AA
intervention is close to 1 (Figure \ref{figTFTnoPS1}).

\begin{table}
\tablewidth=270pt
\caption{Firm2's utility \textit{\texttt{U2}} for a market with imperfect
substitutability}
\label{tabps1}
\begin{tabular*}{\tablewidth}{@{\extracolsep{\fill}}lcccc@{}}
\hline
\textbf{\texttt{Firm1}} & \multicolumn{2}{c}{\textbf{Defect}}
& \multicolumn{2}{c@{}}{\textbf{Cooperate}} \\[-4pt]
& \multicolumn{2}{c}{\hrulefill} & \multicolumn{2}{c@{}}{\hrulefill}\\
\textbf{\texttt{Firm2}} & \textbf{Defect} & \textbf{Cooperate}
& \textbf{Defect} & \textbf{Cooperate} \\
\hline
\texttt{U2} & 100 & 50 & 160 & 150 \\
\hline
\end{tabular*}
\end{table}

Furthermore, in contrast to perfect substitutability, accounting
for the presence of entry barriers alone is not sufficient to
modify the optimal decision from cooperate to defect. The main
reason being that when the firms' products are imperfect
substitutes, the set of utilities reflects the fact that the
defect strategy does not correspond to such a strong punishment,
so that a firm can continue to cooperate even if there is high
risk that the game might stop.

\subsubsection{Incomplete information} \label{secincompleteInformation}

Assume that Firm2 has incomplete information about the type of
rival it is going to face. This is a reasonable scenario, as firms
are likely to be uncertain about their rivals' costs and benefits
from cooperation.


%
\begin{table}
\tabcolsep=4pt
\caption{Firm2's expected utility for different values of $\alpha_C$
and $\alpha_D$, without evidence, with evidence $E_1$ and $E_2$,
for likelihood evidence and for the TFT strategy}
\label{tabresults}
\begin{tabular*}{\tablewidth}{@{\extracolsep{\fill}}ld{1.2}cccccc@{}}
\hline
& & \multicolumn{2}{c}{\textbf{Without evidence}}& \multicolumn{2}{c}{\textbf{With
evidence} $\bolds{E_1}$} & \multicolumn{2}{c@{}}{\textbf{With evidence} $\bolds{E_2}$} \\[-4pt]
& & \multicolumn{2}{c}{\hrulefill} & \multicolumn{2}{c}{\hrulefill} & \multicolumn{2}{c@{}}{\hrulefill}\\
$\bolds{\alpha_C}$ & \multicolumn{1}{c}{$\bolds{\alpha_D}$}
& $\bolds{\EV[u(D)]}$ & $\bolds{\EV[u(C)]}$ & $\bolds{\EV
[u(D)|E_1]}$ & $\bolds{\EV[u(C)|E_1]}$ & $\bolds{\EV[u(D)|E_2]}$ & $\bolds{\EV[u(C)|E_2]}$ \\
\hline
1 & 0.25 & 337 & \textit{388} & 329 & \textit{339}& \textit{322} & 298 \\
0.8 & 0.25 & 286 & \textit{316} & \textit{278} & 277 & \textit{271} & 245 \\
0.6 & 0.25 & 238 & \textit{250} & \textit{228} & 219 & \textit{220} & 193 \\
0.4 & 0.25 & \textit{203} & 193 & \textit{190} & 170 & \textit{180} & 152\\
[4pt]
1 & 0.2 & 332 & \textit{388} & 326 & \textit{339} & \textit{321}& 298 \\
0.8 & 0.2 & 281 & \textit{316} & 275 & \textit{277} & \textit{270} & 245 \\
0.6 & 0.2 & 231 & \textit{247} & \textit{225} & 217 & \textit{219} & 193 \\
0.4 & 0.2 & \textit{192} & 188 & \textit{183} & 167 & \textit{177} & 149\\
[4pt]
1 & 0.1 & 321 & \textit{388} & 321 & \textit{339} & \textit{320}& 298 \\
0.8 & 0.1 & 270 & \textit{316} & 270 & \textit{277} & \textit{269} & 245 \\
0.6 & 0.1 & 219 & \textit{243} & \textit{219} & 215 & \textit{218} & 193 \\
0.4 & 0.1 & 172 & \textit{179} & \textit{171} & 159 & \textit{170} & 143 \\
[6pt]
\multicolumn{2}{@{}l}{Likelihood} & 280 & \textit{313} & 273 & \textit{275} &
\textit{268} & 243 \\
\multicolumn{2}{@{}l}{TFT} & 385 & \textit{443} &
390 & \textit{394} & \textit{395} & 353 \\
\hline
\end{tabular*}
\end{table}

Table \ref{tabresults} shows Firm2's expected utility in case of
perfect substitutability (based on\vadjust{\goodbreak} Firm2's utility given in Table
\ref{tabps}) for different probability values of $\alpha_C$ and
$\alpha_D$ (nodes \texttt{alpha\_C} and \texttt{alpha\_D} in Figure
\ref{figglobal}). Three types of information about the relevant market
are considered: no evidence, evidence $E_1=\{\mbox{\texttt{Post}
\texttt{Market} \texttt{Share}}\ge40\%$, \texttt{Entry}
\texttt{Barriers}${}={}$Yes and \texttt{Buyer} \texttt{Power}${}={}$Yes\} and
evidence $E_2$\mbox{${}={}$}\{\texttt{Entry} \texttt{Barriers}${}={}$Yes and
\texttt{HHI} \texttt{Variation}${}\in{}$[500\mbox{--}1000]\}.
The optimal decision
yielding the highest expected utility for each scenario is italicised.

The second last row of Table \ref{tabresults} gives the results when
inserting a uniform likelihood function for $\alpha_C>0.5$ and
$\alpha_D<0.5$. In this case, Firm2's optimal decision is to
cooperate under no evidence and $E_1$. Whereas, for $E_2$, when
the probability of AA intervention is close to one,
$\EV[u(D)|E_2]> \EV[u(C)|E_2]$, so Firm2's optimal decision is to
defect. These results coincide with those obtained using the TFT
strategy shown in the last row of Table \ref{tabresults}. Recall that
the TFT strategy corresponds to setting $\alpha_C=1$ and
$\alpha_D=0$ in all \textbf{Duopoly} instances.

Now, suppose Firm2 believes that its rival cooperates---with
probability $\alpha_C=0.8$---if Firm2 cooperates; and cooperates---with
probability $\alpha_D=0.25$---even if Firm2 defects. This is
implemented in the network inserting and propagating evidence
\texttt{alpha\_C}${}={}$0.8 and \texttt{alpha\_D}${}={}$0.25 in each
\textbf{Duopoly} instance. As we can see in Table \ref{tabresults},
Firm2's expected utility to cooperate, $\EV[u(C)]=316$, is greater than
to defect, $\EV[u(D)]=286$. Introducing evidence $E_1$ in
\textbf{AA\_1}, the two decisions become almost utility equivalent.
Whereas, under the TFT strategy, $E_1$ yields an optimal decision to
cooperate $\EV[u(C)|E_1]=394$, whereas $\EV[u(D)|E_1]= 390$.

Recall that when information about the relevant market is not
taken into account, the probability of AA intervention is 0.0189.
If the probability that Firm1 cooperates when Firm2 defects is
very small ($\alpha_D=0.1$), then its optimal decision is to
cooperate, even for small values of $\alpha_C$. On the other hand,
when $\alpha_D\geq0.2$, defecting is Firm2's best choice for
$\alpha_C=0.4$, yielding a different behaviour from that obtained
using the TFT strategy. However, using evidence $E_1$, when the
probability of AA intervention is 0.514, $\EV[u(D)|E_1] >
\EV[u(C)|E_1]$ even when Firm1 is slightly altruistic,
$\alpha_D\leq0.2$ and $\alpha_C\leq0.6$. Furthermore, if
$\alpha_D=0.25$, then $\EV[u(D)|E_1] > \EV[u(C)|E_1]$ also for
$\alpha_C\leq0.8$. If the TFT strategy is adopted, Firm2
optimally cooperates both under no evidence and $E_1$, whereas
for $E_2$ the associated probability of AA intervention is very
large, so that Firm2's optimal decision is to defect for all
values of $\alpha_C$ and $\alpha_D$ considered here.

While the examples shown here are merely illustrative, the number
of questions and different strategies that can be analysed is
clearly huge and increases with the number of stages considered.

\section{Conclusion}\label{secconclusions}



When the antitrust authority starts an investigation, the two
potentially merging firms are likely to represent a relevant share
of the market, hence, they might affect the price of the goods
traded. In contrast, the decisions of other firms inside the
market, but outside the merged entity, can be assumed to be
irrelevant. In circumstances such as these, a PD duopoly model is
a reasonable representation.



From an economic perspective, the methodology we present can be seen as
a useful decision support system. It models and integrates the different
uncertainty sources deriving from a rival competitor and from the
economic environment. Furthermore, the model can be updated as we
consider new
cases, changes in market conditions or new antitrust regulations. The
emphasis in this paper is to show the potentiality of OOBNs in the analysis
of duopoly markets with external uncertainty. For the sake of
simplicity, the duopoly is represented by a rather naive game theoretic
model; in
future studies we wish to implement a more complex interaction model
between firms.

As is standard in industrial organization, the firm is seen as a single
decision making unit; generalisations of our OOBN to model firms' internal
organization could also be considered. Indeed, a firm's top and middle
management may have different objectives from its owner. An appropriate BN
could be built to model these interrelationships and incorporate them
into a more general OOBN model.
This would yield a more complete and realistic picture of firms'
cooperative behaviour. We hope to develop this and other aspects in the future.


%
%
%

\section*{Acknowledgements}

We are indebted to Carlo Cazzola, Mauro La Noce and Valerio Ruocco of
the Italian Antitrust Authority for providing the data. We thank the
referee and Associate Editor for their insightful comments. We also
thank Mario Tirelli and Filippo Calciano for suggestions on a previous
version.



\printaddresses

\end{document}